\documentclass[conference]{IEEEtran}
\IEEEoverridecommandlockouts

\usepackage{amsmath,amsfonts}
\usepackage{algorithmic}
\usepackage{algorithm}
\usepackage{array}
\usepackage[caption=false,font=footnotesize,labelfont=rm,textfont=rm]{subfig}
\usepackage{textcomp}
\usepackage{stfloats}
\usepackage{url}
\usepackage{verbatim}
\usepackage{graphicx}
\usepackage{cite}
\usepackage{enumitem}
\usepackage[breaklinks,colorlinks, linkcolor=red, citecolor=blue, urlcolor=black]{hyperref}

\usepackage[utf8]{inputenc} 
\usepackage[T1]{fontenc}    
\usepackage{hyperref}       
\usepackage{url}            
\usepackage{booktabs}       
\usepackage{amsfonts}       
\usepackage{nicefrac}       
\usepackage{microtype}      
\usepackage{xcolor}         
\usepackage{multirow}
\usepackage{graphicx}
\usepackage{epstopdf}
\usepackage{amsmath}
\usepackage{amssymb}
\usepackage{utfsym}
\usepackage{pifont}
\usepackage{bm}
\usepackage{float}
\usepackage{subfig}
\usepackage{wrapfig}
\usepackage[normalem]{ulem}
\newcommand{\ie}{{\emph{i.e.}}}

\newcommand{\etal}{{\emph{et al.}}}

\newcommand{\cmark}{\ding{51}}%
\newcommand{\xmark}{\ding{55}}

\usepackage{marvosym}

\title{Self-Supervised Learning for Detecting \\AI-Generated Faces as Anomalies}
\author{\IEEEauthorblockN{Mian Zou}
\IEEEauthorblockA{
\textit{City University of Hong Kong}\\
Hong Kong, Hong Kong \\
mianzou2-c@my.cityu.edu.hk}
\and
\IEEEauthorblockN{Baosheng Yu}
\IEEEauthorblockA{
\textit{Nanyang Technological University}\\
Singapore \\
baosheng.yu@ntu.edu.sg}
\and
\IEEEauthorblockN{Yibing Zhan}
\IEEEauthorblockA{\textit{JD Explore Academy} \\
Beijing, China \\
zhanyibing@jd.com}
\and
\IEEEauthorblockN{Kede Ma\textsuperscript{\textrm{\Letter}}\thanks{\textsuperscript{\textrm{\Letter}}Corresponding author.}}
\IEEEauthorblockA{
\textit{City University of Hong Kong}\\
Hong Kong, Hong Kong \\
kede.ma@cityu.edu.hk}
}
\begin{document}
%
\maketitle
\begin{abstract}
The detection of AI-generated faces is commonly approached as a binary classification task.
Nevertheless, the resulting detectors frequently struggle to adapt to novel AI face generators, which evolve rapidly. In this paper, we describe an anomaly detection method for AI-generated faces by leveraging self-supervised learning of camera-intrinsic and face-specific features purely from photographic face images. The success of our method lies in designing a pretext task that trains a feature extractor to rank four ordinal exchangeable image file format (EXIF) tags and classify artificially manipulated face images. Subsequently, we model the learned feature distribution of photographic face images using a Gaussian mixture model. Faces with low likelihoods are flagged as AI-generated. Both quantitative and qualitative experiments validate the effectiveness of our method. Our code is available at \url{https://github.com/MZMMSEC/AIGFD_EXIF.git}.

\end{abstract}
\begin{IEEEkeywords}
AI-generated face detection, anomaly detection, self-supervised learning.
\end{IEEEkeywords}

\section{Introduction}
\label{sec: intro}
AI-generated faces, produced using techniques such as generative adversarial networks (GANs)~\cite{karras2020analyzing, esser2021taming} and diffusion models~\cite{rombach2022high, song2021denoising, yu2023freedom, lee2023aligning, Midjourney, podell2024sdxl}, have become nearly indistinguishable from those captured by digital cameras~\cite{nightingale2022ai}. This raises concerns about their potential misuse, making their detection crucial for countering misinformation, preserving multimedia integrity, and ensuring trust in sensitive visual data.

Traditional methods for detecting AI-generated faces typically rely on binary classification, in which detectors are trained to differentiate between photographic and AI-generated images~\cite{johnson2006exposing, matern2019exploiting, liu2020global, yu2019attributing, gragnaniello2021gan, wang2023dire,durall2020watch, corvi2023intriguing, luo2021generalizing, dong2022think, frank2020leveraging}. However, such methods often lack generalizability to newly emerging AI face generators with distinct design philosophies, higher-quality training data, and better optimization pipelines. As a result, existing detectors that are overly tuned to specific generative models may quickly become obsolete when faced with unfamiliar generators.

To address this generalization challenge, an alternative strategy treats AI-generated face images as anomalies within the distribution of photographic face images, a concept known as anomaly detection or one-class classification in machine learning~\cite{hawkins1980identification}. Successful implementations of this strategy include exploiting physiological cues like pupil shapes~\cite{guo2022icassp_eyes}, head poses~\cite{yang2019exposing}, and corneal specular  highlights~\cite{hu2021exposing} to expose physical and/or biological irregularities. However, hand-engineered features might have limited utility against photorealistic AI-generated faces. 
Recently, Ojha \etal~\cite{ojha2023towards} demonstrated the potential of pre-trained CLIP features~\cite{radford2021CLIP} for detecting AI-generated images, though CLIP's semantic focus may not be optimal for media forensics.

In this study, we advance the anomaly detection of AI-generated faces through self-supervised representation learning.
Inspired by prior work on exchangeable image file format (EXIF) data for image splicing detection~\cite{huh2018fighting,zheng2023exif}, we design a pretext task that involves ranking~\cite{burges2005learning} four ordinal, EXIF tags (\ie, \texttt{aperture}, \texttt{exposure time}, \texttt{focal length}, and \texttt{ISO speed}) and classifying artificially manipulated face images. Optimized using an equally weighted sum of fidelity losses across these five subtasks~\cite{tsai2007frank, zou2024sjedd}, the learned camera-intrinsic and face-specific features enable anomaly detection of AI-generated faces through a Gaussian mixture model (GMM)~\cite{mclachlan1988mixture}. Extensive experiments on nine state-of-the-art generative models~\cite{karras2020analyzing, esser2021taming, rombach2022high, song2021denoising, yu2023freedom, lee2023aligning, Midjourney, podell2024sdxl} confirm the effectiveness of our method in detecting AI-generated faces.

\begin{figure*}[t]
  \centering
  \subfloat[Self-Supervised Representation Learning]{\includegraphics[width=\linewidth]{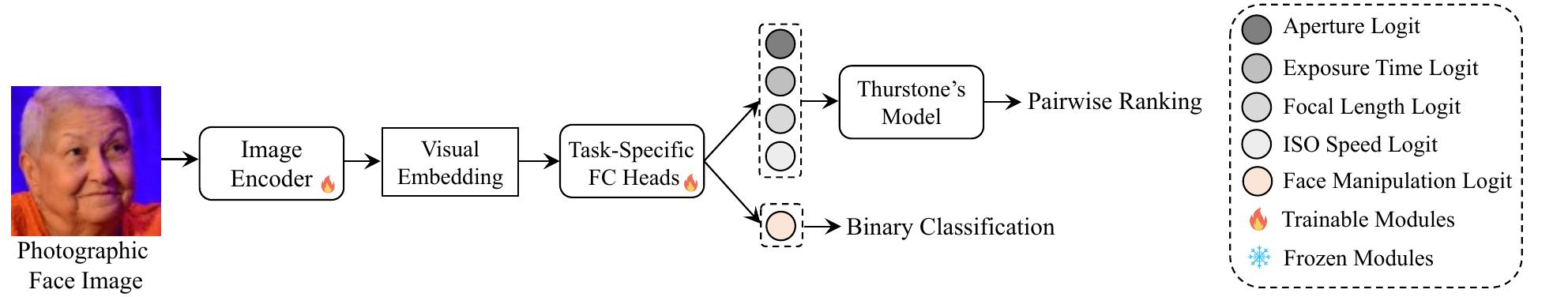}}
  \\
  \subfloat[GMM Training]{\includegraphics[width=0.45\linewidth]{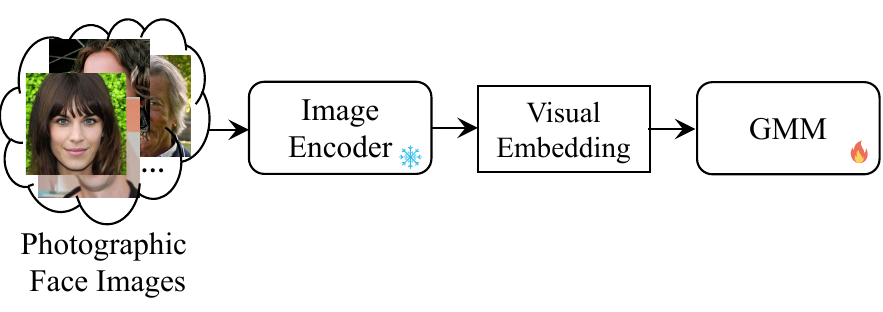}}
  \subfloat[GMM Testing]{\includegraphics[width=0.55\linewidth]{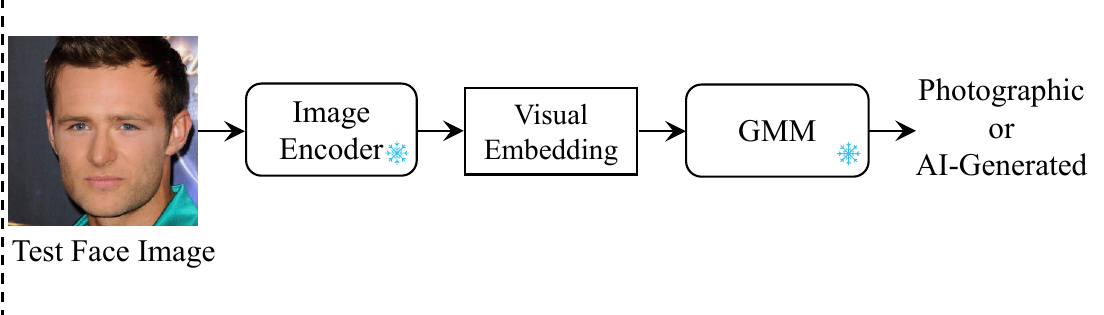}}
  
  \caption{System diagram of the construction of the proposed anomaly detection method for AI-generated faces, including self-supervised representation learning and GMM training. During testing,  faces with low likelihoods are identified as AI-generated.}
  \label{fig: framework}
\end{figure*}

\section{Proposed Method}
\label{sec: method}
In this section, we present in detail the construction of our anomaly detection method for AI-generated faces, including self-supervised representation learning and GMM training. The system diagram is presented in Fig.~\ref{fig: framework}.

\subsection{Self-Supervised Representation Learning}
We design an image feature extractor $\bm f(\cdot;\bm \theta):\mathbb{R}^{H\times W\times 3}\mapsto\mathbb{R}^N$, parameterized by $\bm\theta$, which computes the visual embedding $\bm f(\bm x)$ for a given face image $\bm x\in\mathbb{R}^{H\times W\times 3}$. Here, $H$ and $W$ denote the image height and width, respectively. Additionally, we employ five prediction heads
$g(\cdot;\bm \phi_i):\mathbb{R}^N\mapsto \mathbb{R}$, for $i\in\{1,\ldots,5\}$, collectively parameterized by $\bm \phi$. These heads are used to compute the ``logits'' for four EXIF tag ranking tasks and one face manipulation classification task.

\begin{figure}[]
  \vspace{-.3cm}
  \centering
  \subfloat[]{\includegraphics[width=0.2\linewidth]{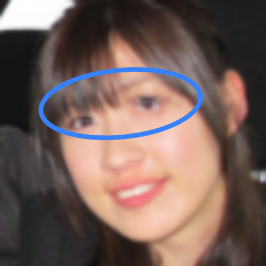}}
  \subfloat[]{\includegraphics[width=0.2\linewidth]{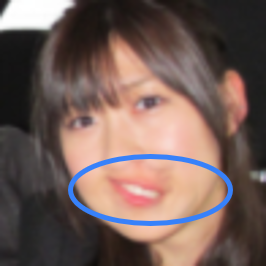}}
  \subfloat[]{\includegraphics[width=0.2\linewidth]{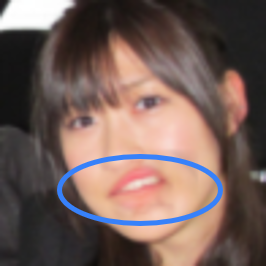}}
  \subfloat[]{\includegraphics[width=0.2\linewidth]{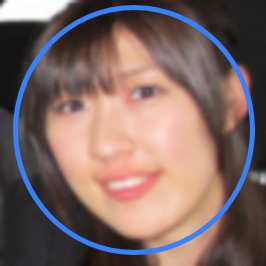}}
  \subfloat[]{\includegraphics[width=0.2\linewidth]{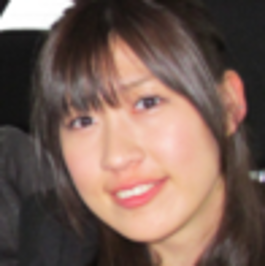}}
  \caption{Visual examples of artificially manipulated face images by \textbf{(a)} horizontal eye flipping, \textbf{(b)} horizontal mouth flipping, \textbf{(c)} vertical mouth flipping, and \textbf{(d)} global affine transformation, respectively. \textbf{(e)} The original face is also shown as the reference.}
  \label{fig: face_negative}
\end{figure}

\noindent\textbf{EXIF Tag Ranking.}
Given a photographic face image pair $(\bm x, \bm y)$, we first derive a binary ground-truth label for the $i$-th EXIF tag:
\begin{align}\label{eq: pair-wise_gts}
     p_i(\bm{x},\bm{y}) = 
\begin{cases} 
 1 & \mbox{if } \mathrm{tag}_i(\bm{x})\ge \mathrm{tag}_i(\bm{y}) \\
      0 & \mbox{otherwise} \end{cases}, \mbox { for } i\in\{1,\ldots, 4\},
\end{align}
where $\mathrm{tag}_i(\cdot)$ outputs the recorded value of the $i$-th tag from the pre-selected EXIF set: \{\texttt{apeture}, \texttt{exposure time}, \texttt{focal length}, \texttt{ISO speed}\}.
Under the Thurstone's model~\cite{thurstone1927law}, we estimate the probability that the $i$-th EXIF tag  of $\bm x$ is larger than that of $\bm y$ as 
\begin{align}\label{eq:thurstone}
\hat{p}(\bm{x}, \bm{y};\bm \theta,\bm \phi_i)= \Phi\left(\frac{g(\bm f(\bm x;\bm\theta);\bm\phi_i)- g(\bm f(\bm y;\bm\theta);\bm\phi_i)}{\sqrt{2}}\right),
\end{align}
where $\Phi(\cdot)$ is the standard normal cumulative distribution function. We then adopt the fidelity loss~\cite{tsai2007frank} for the EXIF tag ranking task:
\begin{align}\label{eq:fidelity}
\ell(\bm{x},\bm{y};\bm \theta,\bm \phi_i)&
= 1 - \sqrt{p_i(\bm{x}, \bm{y})\hat{p}(\bm{x}, \bm{y};\bm \theta,\bm \phi_i)} \nonumber \\ &-\sqrt{(1-p_i(\bm{x}, \bm{y}))(1-\hat{p}(\bm{x}, \bm{y};\bm \theta,\bm \phi_i))}.
\end{align}

\begin{figure*}[]
  \centering
  \includegraphics[width=0.95\linewidth]{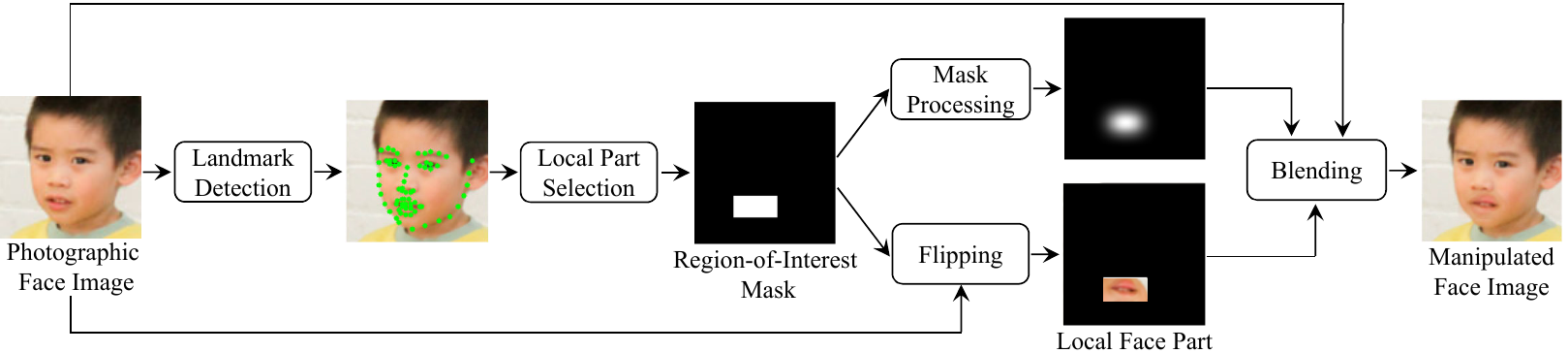}
  \caption{The process of local face part flipping as a form of artificial face manipulation.}
  \label{fig: flipping_negative}
\end{figure*}

\noindent\textbf{Face Manipulation Classification.}
We introduce artificial manipulations to photographic face images via 1) horizontal eye flipping, 2) horizontal mouth flipping, 3) vertical mouth flipping, and 4) global affine transformation, as shown in Fig.~\ref{fig: face_negative}.
The process of local face part flipping is outlined in Fig.~\ref{fig: flipping_negative}. Specifically, for a given photographic face image as input, we first detect its $68$ face landmarks~\cite{bulat2017far} to define a region-of-interest mask. The identified local face part, according to the mask, is then flipped and linearly blended into the original face image. 
For global affine transformation, we utilize the
$\mathtt{PiecewiseAffine}$ API function from the publicly available toolbox---$\mathtt{imgaug}$~\cite{imgaug}.

For consistency, we also employ the fidelity loss for the face manipulation classification task:
\begin{align}\label{eq:cls}
\ell(\bm{x};\bm \theta,\bm \phi_5)&
= 1 - \sqrt{p(\bm{x})\hat{p}(\bm{x};\bm \theta,\bm \phi_5)} \nonumber \\ &-\sqrt{(1-p(\bm{x}))(1-\hat{p}(\bm{x};\bm \theta,\bm \phi_5))},
\end{align}
where $p(\bm{x})=1$ indicates that $\bm x$ has been manipulated and
\begin{align}
    \hat{p}(\bm x;\bm \theta,\bm \phi_5)=\mathtt{Sigmoid}\left(g(\bm f(\bm x;\bm\theta);\bm\phi_5) \right)
\end{align}
is the corresponding estimated probability.

\vspace{3pt}
\noindent\textbf{Overall Loss.}
Given a training minibatch $\mathcal{B}_\mathrm{tr} = \{\bm x^{(m)}\}_{m=1}^M$, we form all possible image pairs and derive the corresponding binary labels using Eq.~\eqref{eq: pair-wise_gts}, which are collectively denoted as $\mathcal{P}_\mathrm{tr}$. Then, the overall loss can be computed by
\begin{align}\label{eq:loss}
\ell(\mathcal{B}_\mathrm{tr};\bm \theta,\bm \phi) = & \frac{1}{\left \vert \mathcal{P}_\mathrm{tr}  \right \vert }\sum_{(\bm x,\bm y)\in \mathcal{P}_\mathrm{tr}} \sum_{i=1}^4 \ell(\bm x,\bm y;\bm \theta, \bm \phi_i) ~ + ~\nonumber\\
& \frac{1}{\left \vert \mathcal{B}_\mathrm{tr}  \right \vert }\sum_{\bm x\in \mathcal{B}_\mathrm{tr}} \ell(\bm x;\bm \theta, \bm \phi_5).
\end{align}

\subsection{GMM Training and Testing}
We approach AI-generated face detection as an anomaly detection problem. To achieve this, we build the probability distribution of the learned features of photographic face images using a GMM with $K$ components:
\begin{align}\label{eq:gmm}
p(\bm z)=\sum_{k=1}^{K}\pi_k \mathcal{N}(\bm z ; \boldsymbol{\mu}_k, \boldsymbol{\Sigma}_k),
\end{align}
where $\bm z=\bm f(\bm x)$ represents the feature vector of $\bm x$. $\pi_k$ is the $k$-th mixing coefficient with the constraint that $\sum_{k=1}^K\pi_k = 1$, and $\boldsymbol{\mu}_k$ and $ \boldsymbol{\Sigma}_k$ are the mean and covariance of the $k$-th Gaussian component. We use the expectation-maximization algorithm for GMM parameter estimation.

During testing, we calculate the log-likelihoods of test image features, flagging those with low likelihoods as AI-generated.

\section{Experiments}
\label{sec: experiments}

\subsection{Experimental Setups}
\noindent\textbf{Datasets.}
We gather $200,000$ face photos, each with four EXIF tags---\texttt{aperture}, \texttt{exposure time}, \texttt{focal length}, and \texttt{ISO speed}---from FDF~\cite{hukkelaas2019deepprivacy} for self-supervised representation learning. 
Following \cite{liu2020global, durall2020watch, cheng2024diffusion, chen2024diffusionface, tan2023learning}, we use $25,000$ face images from the CelebA-HQ dataset~\cite{karras2017progressive} to build the GMM, with additional $5,000$ images reserved for testing. Meanwhile, we collect AI-generated face images from nine state-of-the-art models, including StyleGAN2~\cite{karras2020analyzing}, VQGAN~\cite{esser2021taming}, LDM~\cite{rombach2022high}, DDIM~\cite{song2021denoising}, Stable Diffusion 2.1 (SDv2.1)~\cite{rombach2022high}, FreeDoM~\cite{yu2023freedom}, HPS~\cite{lee2023aligning}, Midjourney~\cite{Midjourney}, and SDXL~\cite{podell2024sdxl}. These images are either sourced from public datasets~\cite{chen2024diffusionface, cheng2024diffusion} or generated on demand~\cite{karras2020analyzing, esser2021taming}.

\begin{table*}[]
\caption{Main results of our method against six representative AI-generated face detectors in terms of Acc and AP. The best results are highlighted in bold, while the second-best results are underlined
}
\label{tab:main_results}
\setlength\tabcolsep{2pt}
\centering
\resizebox{\linewidth}{!}{
\begin{tabular}{lcccccccccccccccccccc}
\toprule
\multirow{2}{*}{Method} & \multicolumn{2}{c}{StyleGAN2} & \multicolumn{2}{c}{VQGAN} & \multicolumn{2}{c}{LDM} & \multicolumn{2}{c}{DDIM} & \multicolumn{2}{l}{SDv2.1} & \multicolumn{2}{c}{FreeDoM} & \multicolumn{2}{c}{HPS} & \multicolumn{2}{c}{Midjourney} & \multicolumn{2}{c}{SDXL} & \multicolumn{2}{c}{Average} \\
\cline{2-21}
                        & Acc           & AP            & Acc         & AP          & Acc        & AP         & Acc         & AP         & Acc          & AP          & Acc          & AP           & Acc        & AP         & Acc            & AP            & Acc         & AP         & Acc          & AP           \\
\hline
GramNet~\cite{liu2020global} & 51.16 & 78.74 & 99.92 & 100.00 & 53.25 & 80.15 & 50.09 & 58.59 & 50.23 & 52.90 & 51.59 & 76.77 & 50.26 & 48.28 & 52.91 & 63.52 & 53.63 & 65.45 & 57.00 & 69.38      \\
FRDM~\cite{luo2021generalizing} & 70.13 & \textbf{99.93} & \textbf{100.00} & \textbf{100.00} & \uline{99.62} & \textbf{100.00} & \textbf{98.26} & \textbf{100.00} & 62.48 & 83.41 & 55.68 & \uline{98.28} & \uline{75.07} & \textbf{96.69} & \textbf{92.51} & \textbf{99.56} & \uline{89.77} & \textbf{99.19} & \uline{82.61} & \textbf{97.45}        \\
RECCE~\cite{Cao_2022_CVPR} & 66.64 & 76.10 & \textbf{100.00} & \textbf{100.00} & 70.91 & 81.69 & 73.10 & 81.15 & 71.62 & 80.71 & 77.52 & 82.80 & 64.14 & \uline{96.40} & 62.19 & 96.28 & 65.29 & 95.67 & 72.38 & 87.87        \\
LGrad~\cite{tan2023learning} & 52.94 & \uline{91.54} & \uline{99.99} & \textbf{100.00} & \textbf{99.80} & \textbf{100.00} & 64.91 & 98.28 & 57.59 & \textbf{93.78} & 66.58 & 49.16 & 60.14 & 96.04 & 76.59 & \uline{98.69} & 74.03 & \uline{98.30} & 72.43 & 91.75        \\
DIRE~\cite{wang2023dire} & \uline{72.48} & 90.34 & 69.81 & 89.82 & 98.92 & \textbf{100.00} & 77.80 & 91.80 & 58.84 & 74.16 & \uline{89.05} & 94.26 & 62.50 & 78.30 & 90.75 & 97.43 & 87.79 & 96.46 & 78.66 & 90.29        \\
UnivFD~\cite{ojha2023towards} & 65.45 & 71.47 & 83.40 & \uline{97.58} & 70.06 & 76.61 & 72.25 & 81.02 & \uline{72.76} & 80.66 & 78.55 & 86.88 & 56.21 & 60.10 & 54.96 & 56.35 & 58.01 & 60.76 & 67.96 & 74.60        \\
\hline
Ours & \textbf{76.88} & 83.94 & 74.59 & 82.85 & 93.83 & \uline{98.67} & \uline{93.63} & \uline{98.67} & \textbf{78.62} & \uline{87.60} & \textbf{95.31} & \textbf{99.86} & \textbf{83.79} & 91.15 & \uline{91.29} & 97.32 & \textbf{91.71} & 97.38 & \textbf{86.63} & \uline{93.01}        \\
\bottomrule
\end{tabular}
}
\end{table*}

\begin{table}[]
\caption{AUC results of GMM-based detectors using different pre-trained features. EXIF-L2R is a degenerate of our method that only learns to rank the four EXIF tags}
\label{tab:pretrained_feature}
\small
\centering
\resizebox{\linewidth}{!}{
\begin{tabular}{lcccccc}
\toprule
Generator  & CLIP & FaRL & EXIF-LAN & EXIF-L2R & Ours  \\
\hline
StyleGAN2  &  33.99    &  34.35    & 69.71    &  67.91  & \textbf{85.69} \\
VQGAN      &  60.20    &  55.91    & 72.41    &  69.99  & \textbf{84.31} \\
LDM        &  55.49    &  47.26    & 85.81    &  83.20  & \textbf{98.66} \\
DDIM       &  82.85    &  85.10    & 84.98    &  87.94  & \textbf{97.96} \\
SDv2.1     &  90.15    &  \textbf{95.24}    & 74.21    & 86.74   & 88.29 \\
FreeDom    &  85.39    &  79.91    & 97.92    &  93.84  & \textbf{99.79} \\
HPS        &  91.22    &  \textbf{93.97}    & 87.33    &  90.30  & 91.92 \\
Midjourney &  92.21    &  89.72    & 91.65    &  94.06  & \textbf{97.07} \\
SDXL       &  93.66    &  94.61    & 93.04    &  95.39  & \textbf{97.23} \\
\hline
Average    &  76.13    &  75.12    & 84.12    &  85.49  & \textbf{93.43} \\
\bottomrule
\end{tabular}
}
\end{table}

\noindent\textbf{Implementation Details.}
Our method utilizes ResNet-50~\cite{he2016deep} as the backbone for feature extraction, resulting in a feature dimension $N$ of $768$. We use the initializations from~\cite{zheng2023exif} and set the input image size to $224\times224\times3$.
During self-supervised representation learning, we minimize the objective in Eq.~\eqref{eq:loss} using Adam~\cite{Kingma2014adam} with a decoupled weight decay of $10^{-4}$ and a minibatch size of $256$, for a total of $20$ epochs. The initial learning rate is set to $10^{-5}$, which follows a cosine annealing schedule~\cite{loshchilov2016sgdr}. 
The number of Gaussian components $K$ in the GMM is set to eight.

\noindent\textbf{Evaluation Metrics.} 
We adopt the detection accuracy (Acc (\%)) and average precision (AP (\%)) as the evaluation metrics. We set a low-likelihood threshold that gives a $5\%$ false alarm rate (\ie, the $5$-th percentile) in the training set to screen AI-generated faces. Additionally, 
we employ the area under the curve (AUC (\%)) to assess the discriminative power of pre-trained features.

\subsection{Main Results}
In line with prior studies~\cite{ojha2023towards,cozzolino2024zed,corvi2023detection}, our experiments focus on evaluating generalizability.
We compare our method with six representative detectors: GramNet~\cite{liu2020global}, FRDM~\cite{luo2021generalizing}, RECCE~\cite{Cao_2022_CVPR}, LGrad~\cite{tan2023learning}, UnivFD~\cite{ojha2023towards}, and DIRE~\cite{wang2023dire}. 
For a fair comparison, we retrain all competing methods on the same datasets (\ie, CelebA-HQ and VQGAN-generated images), adhering strictly to the training protocols and hyperparameter settings described in the original publications. The only exception is DIRE, which is trained on the CelebA-HQ and LDM datasets, as it relies on reconstruction errors of diffusion models.
Table~\ref{tab:main_results} shows the main results, from which we have some key observations. First, despite being trained exclusively on photographic face images, our method archives the highest average accuracy, surpassing the second-best detector by a substantial margin of $4.02 \%$. 
Notably, our method successfully detects AI-generated faces by commercial APIs such as Midjourney~\cite{Midjourney} and SDv2.1~\cite{rombach2022high}.
Second, UnivFD, which utilizes semantic CLIP features~\cite{radford2021CLIP}, exhibits marginal performance, indicating that semantics of photographic and AI-generated faces become increasingly indistinguishable, consistent with the findings in \cite{nightingale2022ai}. Third, frequency-based (\ie, FRDM~\cite{luo2021generalizing}) and gradient-based (\ie, LGrad~\cite{tan2023learning}) methods, trained on GAN-based images, also perform well on diffusion-based images in terms of AP, suggesting a shared presence of low-level artifacts as detection cues. However, their relatively lower accuracies point to potential overfitting issues. Finally, DIRE~\cite{wang2023dire} encounters challenges in generalizing to the same family of diffusion models, emphasizing the significant impact of the evolution of generative models on detection accuracy.

\begin{table}[]
\caption{Ablation on the pretext task for self-supervised representation learning}
\label{tab:ablations}
\small
\centering
\begin{tabular}{cccc}
\toprule
\begin{tabular}[c]{@{}c@{}}EXIF Tag \\ Ranking\end{tabular}   & \begin{tabular}[c]{@{}c@{}}Face  Manipulation\\ Classification\end{tabular} & Mean Acc & Mean AP \\
\hline
\cmark & \xmark & 69.35 & 86.44 \\
\xmark & \cmark & 77.67 & 84.23 \\
\hline
\cmark & \cmark & \textbf{86.63} & \textbf{93.01} \\
\bottomrule
\end{tabular}
\end{table}

\begin{figure}[]
  \centering
  \includegraphics[width=0.7\linewidth]{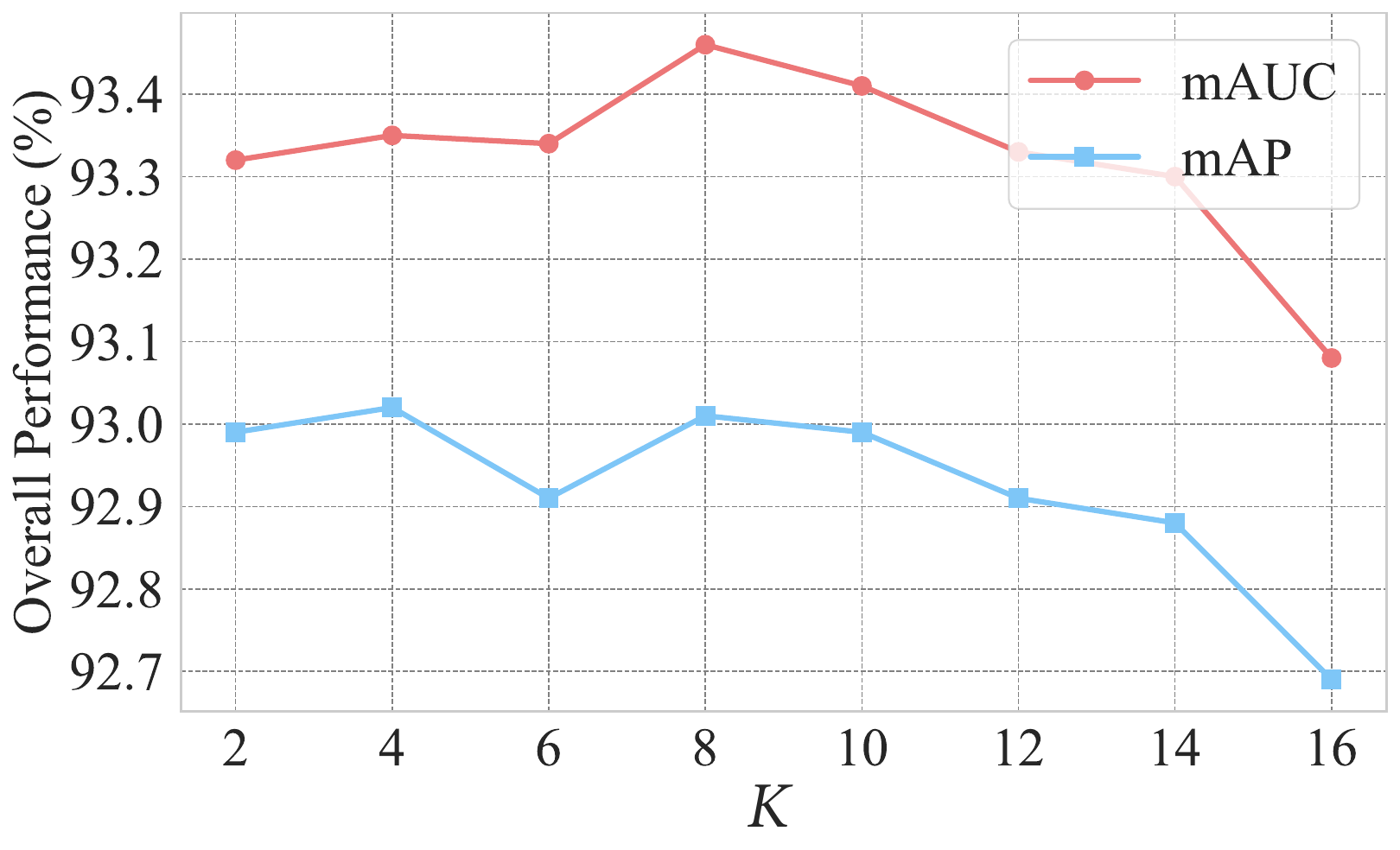}
  \caption{Ablation on the number of Gaussian components of the GMM.}
  \label{fig: gmm-k}
\end{figure}

\subsection{Comparisons with Other Pre-Trained Features}
We compare our proposed self-supervised features against four alternatives by 1) CLIP ~\cite{radford2021CLIP}, 2) FaRL~\cite{zheng2022general}, 3) EXIF-as-language (EXIF-LAN)~\cite{zheng2023exif}, and 4) a degenerate of our method that only learns to rank the four EXIF tags (EXIF-L2R), within the same GMM framework. CLIP is pre-trained contrastively on massive image-text pairs to capture high-level image semantics, while FaRL and EXIF-LAN can be seen as variants of CLIP. FaRL is pre-trained solely on face images, and EXIF-LAN aligns images with corresponding EXIF tags.
As shown in Table~\ref{tab:pretrained_feature}, our features outperform all counterparts, validating
the usefulness of the proposed self-supervised learning method in improving feature discriminability.
In contrast, semantics-oriented features (\ie, by CLIP and FaRL) fall short in detecting AI-generated faces.
In addition, the superiority of EXIF-L2R over EXIF-LAN verifies the importance of capturing more fine-grained ordinal information of EXIF tags in learning  EXIF-induced features.

\subsection{Further Analysis}
\noindent\textbf{Ablation Studies.} 
We first analyze two degenerates of our pretext task: 1) minimizing only Eq.~\eqref{eq:fidelity} and 2)  minimizing only Eq.~\eqref{eq:cls} in self-supervised representation learning. From the results in Table~\ref{tab:ablations}, it is clear that the full pretext task results in the best performance, thereby confirming its design rationale. Next, 
we tune the number of Gaussian components (\ie, $K$ in Eq.~\eqref{eq:gmm}) on a separate validate set to see its variability in detection performance. Fig.~\ref{fig: gmm-k} shows fairly stable performance across different values of $K$, with the best results at $K=8$.

\begin{figure}[]
  \centering
  \includegraphics[width=\linewidth]{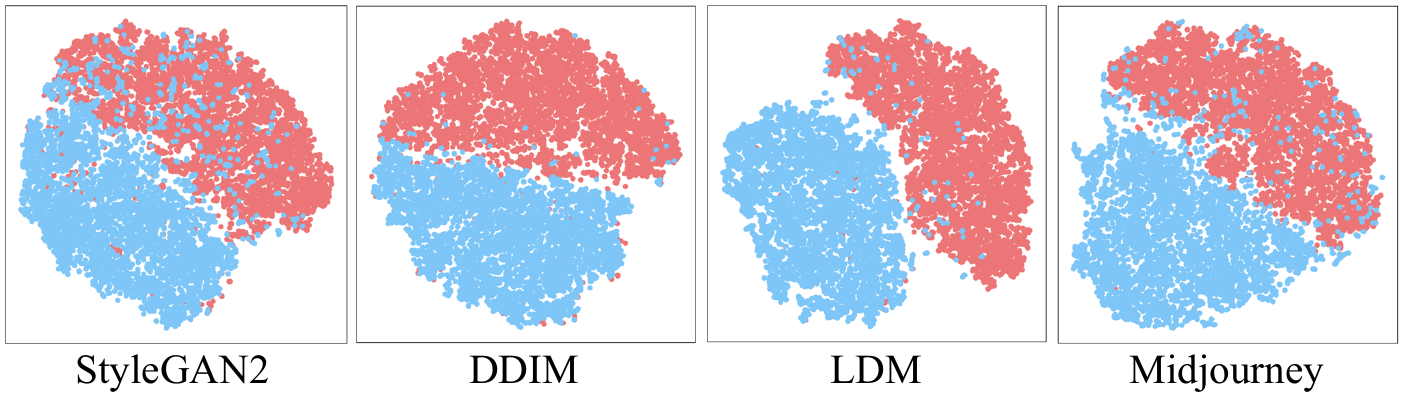}
  \caption{t-SNE embeddings~\cite{van2008visualizing} of self-supervised features for photographic (in red) and AI-generated (in blue) face images.}
  \label{fig: tsne_feature}
\end{figure}

\noindent\textbf{Visualizations.} 
Fig.~\ref{fig: tsne_feature} illustrates 2D embeddings of pre-trained features for photographic and AI-generated faces using t-SNE~\cite{van2008visualizing}, highlighting a clear separation between the two classes. This further validates the effectiveness of our self-supervised representation learning method in capturing unique characteristics of photographic face images.

\section{Conclusion and Discussion}
\label{sec: conclusion}
We have introduced an AI-generated face detector based on self-supervised anomaly detection.
Central to our method is the design of a pretext task that encourages learning camera intrinsic and face-specific features from photographic face images. 
Experimental results demonstrate the promise of our self-supervised features in detecting AI-generated faces.

The present study is limited to detecting AI-generated faces. Future work could expand the proposed idea to detect a broader range of AI-generated content. Moreover, exploring the joint optimization of self-supervised representation learning and anomaly detection, potentially in a bilevel framework~\cite{dempe2002foundations}, could be a promising direction.
Additionally, leveraging self-supervised features as guidance~\cite{li2017learning, haliassos2022leveraging} could potentially improve the training of a binary classifier for AI-generated faces~\cite{liu2020global, wang2019cnngenerated,luo2021generalizing, tan2023learning, wang2023dire}.

\section*{Acknowledgements}
This work was supported in part by the Hong Kong RGC General Research Fund (11220224), the CityU Strategic Research Grants (7005848 and 7005983), and an Industry Gift Fund (9229179).

\begingroup
    \bibliographystyle{IEEEbib}
    \bibliography{DeepFake-25icassp}
\endgroup

\end{document}